\documentclass[11pt,letterpaper]{article}
\usepackage{CJK}
\usepackage{emnlp2016}
\usepackage{times}
\usepackage{latexsym}
\usepackage{amsmath,amsfonts,amssymb}
\usepackage{enumerate}
\usepackage{color}
\usepackage{mdframed}
\usepackage{multirow}
\usepackage{multicol}
\usepackage{textcomp}
\usepackage{subcaption}
\usepackage[dvipdfmx]{graphicx}
\usepackage{mathtools}
\usepackage{bm}
\usepackage{url}
\usepackage{amsmath}
\emnlpfinalcopy

\def\emnlppaperid{799}

\DeclareMathOperator*{\argmin}{argmin}

\usepackage{xcolor}

\title{Incorporating Discrete Translation Lexicons \\ into Neural Machine Translation}

\author{Philip Arthur$^*$, Graham Neubig$^{*\dag}$, Satoshi Nakamura$^*$\\
  {$^*$ Graduate School of Information Science, Nara Institute of Science and Technology}\\
  {$^\dag$ Language Technologies Institute, Carnegie Mellon University}\\
  {\tt{philip.arthur.om0@is.naist.jp} \tt{gneubig@cs.cmu.edu} \tt{s-nakamura@is.naist.jp}}}

\date{\today}

\begin{document}
\begin{CJK}{UTF8}{min}

\maketitle

\begin{abstract}
	Neural machine translation (NMT) often makes mistakes in translating low-frequency content words that are essential to understanding the meaning of the sentence.
    We propose a method to alleviate this problem by augmenting NMT systems with discrete translation lexicons that efficiently encode translations of these low-frequency words.
    We describe a method to calculate the lexicon probability of the next word in the translation candidate by using the attention vector of the NMT model to select which source word lexical probabilities the model should focus on. 
    We test two methods to combine this probability with the standard NMT probability: (1) using it as a bias, and (2) linear interpolation.
    Experiments on two corpora show an improvement of 2.0-2.3 BLEU and 0.13-0.44 NIST score, and faster convergence time.\footnote{Tools to replicate our experiments can be found at {http://isw3.naist.jp/\texttildelow philip-a/emnlp2016/index.html}}
\end{abstract}

\section{Introduction}
Neural machine translation (NMT, \S\ref{sec:nmt}; \newcite{kalchbrenner13emnlp}, \newcite{sutskever14nips}) is a variant of statistical machine translation (SMT; \newcite{brown93cl}), using neural networks.
NMT has recently gained popularity due to its ability to model the translation process end-to-end using a single probabilistic model, and for its state-of-the-art performance on several language pairs \cite{luong15emnlp,sennrich16acl}.

One feature of NMT systems is that they treat each word in the vocabulary as a vector of continuous-valued numbers.
This is in contrast to more traditional SMT methods such as phrase-based machine translation (PBMT; \newcite{koehn03phrasebased}), which represent translations as discrete pairs of word strings in the source and target languages.
The use of continuous representations is a major advantage, allowing NMT to share statistical power between similar words (e.g. ``dog'' and ``cat'') or contexts (e.g. ``this is'' and ``that is''). 
However, this property also has a drawback in that NMT systems often mistranslate into words that seem natural in the context, but do not reflect the content of the source sentence.
For example, Figure \ref{fig:intro_mistake} is a sentence from our data where the NMT system mistakenly translated ``Tunisia'' into the word for ``Norway.''
This variety of error is particularly serious because the content words that are often mistranslated by NMT are also the words that play a key role in determining the whole meaning of the sentence.

\begin{figure}[t]
\centering
        \begin{tabular}{l l}
        \textbf{Input:}   & I come from \underline{Tunisia}. \\
        \textbf{Reference:}   & \underline{チュニジア} の 出身です。 \\
                        & \small{\underline{Chunisia} no shusshindesu.} \\
                        & \textit{(I'm from Tunisia.)} \\
        \textbf{System:}   & \underline{ノルウェー} の 出身です。 \\
                        & \small{\underline{Noruue-} no shusshindesu.} \\
                        & \textit{(I'm from Norway.)} \\
        \end{tabular}
    \caption{An example of a mistake made by NMT on low-frequency content words.}
    \label{fig:intro_mistake}
\end{figure}

In contrast, PBMT and other traditional SMT methods tend to rarely make this kind of mistake.
This is because they base their translations on discrete phrase mappings, which ensure that source words will be translated into a target word that has been observed as a translation at least once in the training data.
In addition, because the discrete mappings are memorized explicitly, they can be learned efficiently from as little as a single instance (barring errors in word alignments).
Thus we hypothesize that if we can incorporate a similar variety of information into NMT, this has the potential to alleviate problems with the previously mentioned fatal errors on low-frequency words.

In this paper, we propose a simple, yet effective method to incorporate discrete, probabilistic lexicons as an additional information source in NMT (\S\ref{sec:using}).
First we demonstrate how to transform lexical translation probabilities (\S\ref{sec:construct}) into a predictive probability for the next word by utilizing attention vectors from attentional NMT models \cite{bahdanau15iclr}.
We then describe methods to incorporate this probability into NMT, either through linear interpolation with the NMT probabilities (\S\ref{sec:linear}) or as the bias to the NMT predictive distribution (\S\ref{sec:bias}).
We construct these lexicon probabilities by using traditional word alignment methods on the training data (\S\ref{sec:automatic}), other external parallel data resources such as a handmade dictionary (\S\ref{sec:handmade}), or using a hybrid between the two (\S\ref{sec:hybrid}).

We perform experiments (\S\ref{sec:experiment}) on two English-Japanese translation corpora to evaluate the method's utility in improving translation accuracy and reducing the time required for training.

\section{Neural Machine Translation}
\label{sec:nmt}
The goal of machine translation is to translate a sequence of source words $F=f^{\lvert F \rvert}_1$ into a sequence of target words $E=e^{\lvert E \rvert}_1$.
These words belong to the source vocabulary $V_f$, and the target vocabulary $V_e$ respectively.
NMT performs this translation by calculating the conditional probability $p_m(e_i | F, e^{i-1}_1)$ of the $i$th target word $e_i$ based on the source $F$ and the preceding target words $e^{i-1}_1$.
This is done by encoding the context $\langle F, e^{i-1}_1 \rangle$ a fixed-width vector $\bm{\eta}_i$, and calculating the probability as follows:
\begin{equation}
    \label{eq:softmax}
    p_m(e_i| F, e_1^{i-1}) = \text{softmax}(W_{s} \bm{\eta}_i + \bm{b}_{s}),
\end{equation}
where $W_{s}$ and $\bm{b}_{s}$ are respectively weight matrix and bias vector parameters.

The exact variety of the NMT model depends on how we calculate $\bm{\eta}_i$ used as input.
While there are many methods to perform this modeling, we opt to use attentional models \cite{bahdanau15iclr}, which focus on particular words in the source sentence when calculating the probability of $e_i$.
These models represent the current state of the art in NMT, and are also convenient for use in our proposed method.
Specifically, we use the method of \newcite{luong15emnlp}, which we describe briefly here and refer readers to the original paper for details.

First, an \textit{encoder} converts the source sentence $F$ into a matrix $R$ where each column represents a single word in the input sentence as a continuous vector.
This representation is generated using a bidirectional encoder
\begin{eqnarray*}
    \overrightarrow{\bm{r}}_j = \text{enc}( \text{embed}(f_j), \overrightarrow{\bm{r}}_{j-1} ) \\
    \overleftarrow{\bm{r}}_j = \text{enc}( \text{embed}(f_j), \overleftarrow{\bm{r}}_{j+1} ) \\
    \bm{r}_j = [\overleftarrow{\bm{r}}_j; \overrightarrow{\bm{r}}_j].
\end{eqnarray*}
Here the $\text{embed}(\cdot)$ function maps the words into a representation \cite{bengio03jmlr}, and $\text{enc}(\cdot)$ is a stacking long short term memory (LSTM) neural network \cite{hochreiter97nc,gers00nc,sutskever14nips}.
Finally we concatenate the two vectors $\overrightarrow{\bm{r}}_j$ and $\overleftarrow{\bm{r}}_j$ into a bidirectional representation $\bm{r}_j$.
These vectors are further concatenated into the matrix $R$ where the $j$th column corresponds to $\bm{r}_j$.

Next, we generate the output one word at a time while referencing this encoded input sentence and tracking progress with a \textit{decoder} LSTM.
The decoder's hidden state $\bm{h}_i$ is a fixed-length continuous vector representing the previous target words $e^{i-1}_1$, initialized as $\bm{h}_0 = \bm{0}$.
Based on this $\bm{h}_i$, we calculate a similarity vector $\bm{\alpha}_i$, with each element equal to
\begin{equation}
\alpha_{i,j} = \text{sim}(\bm{h}_i,\bm{r}_j).
\end{equation}
$\text{sim}(\cdot)$ can be an arbitrary similarity function, which we set to the dot product, following \newcite{luong15emnlp}.
We then normalize this into an \textit{attention} vector, which weights the amount of focus that we put on each word in the source sentence
\begin{equation}
\label{eq:attention}
\bm{a}_i = \text{softmax}(\bm{\alpha}_i).
\end{equation}
This attention vector is then used to weight the encoded representation $R$ to create a context vector $\bm{c_i}$ for the current time step
\begin{equation*}
    \bm{c} = R \bm{a}.
\end{equation*}
Finally, we create $\bm{\eta}_i$ by concatenating the previous hidden state $\bm{h}_{i-1}$ with the context vector, and performing an affine transform
\begin{equation*}
    \bm{\eta}_i = W_{\eta}[\bm{h}_{i-1}; \bm{c}_i] + b_{\eta},
\end{equation*}

Once we have this representation of the current state, we can calculate $p_m(e_i | F, e^{i-1}_1)$ according to Equation (\ref{eq:softmax}).
The next word $e_i$ is chosen according to this probability, and we update the hidden state by inputting the chosen word into the decoder LSTM
\begin{equation}
    \bm{h}_i = \text{enc}( \text{embed}(e_i), \bm{h}_{i-1} ).
\end{equation}

If we define all the parameters in this model as $\theta$, we can then train the model by minimizing the negative log-likelihood of the training data
\begin{equation*}
    \hat{\theta} = \argmin_{\theta} \sum_{\langle F,~E \rangle} \sum_{i} -\log(p_m(e_i | F, e^{i-1}_1; \theta)).
\end{equation*}

\section{Integrating Lexicons into NMT}
\label{sec:using}
In \S\ref{sec:nmt} we described how traditional NMT models calculate the probability of the next target word $p_m(e_i | e_1^{i-1}, F)$.
Our goal in this paper is to improve the accuracy of this probability estimate by incorporating information from discrete probabilistic lexicons.
We assume that we have a lexicon that, given a source word $f$, assigns a probability $p_l(e | f)$ to target word $e$.
For a source word $f$, this probability will generally be non-zero for a small number of translation candidates, and zero for the majority of words in $V_E$.
In this section, we first describe how we incorporate these probabilities into NMT, and explain how we actually obtain the $p_l(e|f)$ probabilities in \S\ref{sec:type}.

\subsection{Converting Lexicon Probabilities into Conditioned Predictive Proabilities}
\label{sec:construct}

First, we need to convert lexical probabilities $p_l(e | f)$ for the individual words in the source sentence $F$ to a form that can be used together with $p_m(e_i | e_1^{i-1}, F)$.
Given input sentence $F$, we can construct a matrix in which each column corresponds to a word in the input sentence, each row corresponds to a word in the $V_E$, and the entry corresponds to the appropriate lexical probability:
\small
\begin{equation*}
    L_F = 
        \begin{bmatrix}
            ~p_{l}(e=1 | f_1)~ & \cdots & p_{l}(e=1|f_{\lvert F \rvert})\\
            \vdots & \ddots & \vdots \\
            ~p_{l}(e={\lvert V_e \lvert} | f_1) & \cdots & p_{l}(e={\lvert V_e \lvert}| f_{\lvert F \rvert})~
        \end{bmatrix}.
\end{equation*}
\normalsize
This matrix can be precomputed during the encoding stage because it only requires information about the source sentence $F$.

Next we convert this matrix into a predictive probability over the next word: $p_l(e_i|F, e^{i-1}_1)$.
To do so we use the alignment probability $\bm{a}$ from Equation (\ref{eq:attention}) to weight each column of the $L_F$ matrix:
\small
\begin{multline*}
    p_{l}(e_i|F, e^{i-1}_1) = L_F \bm{a}_i  = \\ 
        \begin{bmatrix}
            p_{l}(e=1|f_1) & \cdots & p_{lex}(e=1|f_{\lvert F \rvert}) \\
            \vdots   & \ddots & \vdots \\
            p_{l}(e={V_e}|f_1) & \cdots & p_{lex}(e={V_e}|f_{\lvert F \rvert})
        \end{bmatrix}
        \begin{bmatrix}
            a_{i,1} \\
            \vdots \\
            a_{i,\lvert F \rvert}
        \end{bmatrix}.
\end{multline*}
\normalsize
This calculation is similar to the way how attentional models calculate the context vector $\bm{c}_i$, but over a vector representing the probabilities of the target vocabulary, instead of the distributed representations of the source words.
The process of involving $\bm{a}_i$ is important because at every time step $i$, the lexical probability $p_l(e_i | e_1^{i-1}, F)$ will be influenced by different source words.

\subsection{Combining Predictive Probabilities}
After calculating the lexicon predictive probability $p_l(e_i | e_1^{i-1}, F)$, next we need to integrate this probability with the NMT model probability $p_m(e_i | e_1^{i-1}, F)$.
To do so, we examine two methods: (1) adding it as a bias, and (2) linear interpolation.

\subsubsection{Model Bias}
\label{sec:bias}
In our first \texttt{bias} method, we use $p_l(\cdot)$ to bias the probability distribution calculated by the vanilla NMT model.
Specifically, we add a small constant $\epsilon$ to $p_{l}(\cdot)$, take the logarithm, and add this adjusted log probability to the input of the softmax as follows:
\begin{multline*}
    p_b(e_i| F, e_1^{i-1}) = \text{softmax}(W_{s}\bm{\eta}_i + b_{s} + \\ \log(p_{l}(e_i | F, e^{i-1}_1) + \epsilon)).
\end{multline*}
We take the logarithm of $p_{l}(\cdot)$ so that the values will still be in the probability domain after the softmax is calculated, and add the hyper-parameter $\epsilon$ to prevent zero probabilities from becoming $-\infty$ after taking the log.
When $\epsilon$ is small, the model will be more heavily biased towards using the lexicon, and when $\epsilon$ is larger the lexicon probabilities will be given less weight.
We use $\epsilon=0.001$ for this paper.


\subsubsection{Linear Interpolation}
\label{sec:linear}
We also attempt to incorporate the two probabilities through linear interpolation between the standard NMT probability model probability $p_m(\cdot)$ and the lexicon probability $p_l(\cdot)$.
We will call this the {\tt linear} method, and define it as follows:

\small
\begin{multline*}
    p_o(e_i | F, e^{i-1}_1) = \\
    \begin{bmatrix}
        p_{l}(e_i=1 | F, e^{i-1}_1) && p_{m}(e=1 | F, e^{i-1}_1)\\
        \vdots && \vdots \\
        p_{l}(e_i=\lvert V_e \rvert | F, e^{i-1}_1) && p_{m}(e=\lvert V_e \rvert | F, e^{i-1}_1)
    \end{bmatrix}
    \begin{bmatrix}
        \lambda \\
        1 - \lambda
    \end{bmatrix},
\end{multline*}
\normalsize
where $\lambda$ is an interpolation coefficient that is the result of the sigmoid function $\lambda = \text{sig}(x) = \frac{1}{1+e^{-x}}$.
$x$ is a learnable parameter, and the sigmoid function ensures that the final interpolation level falls between 0 and 1.
We choose $x=0$ ($\lambda=0.5$) at the beginning of training.

This notation is partly inspired by \newcite{allamanis16icml} and \newcite{gu16acl} who use linear interpolation to merge a standard attentional model with a ``copy'' operator that copies a source word as-is into the target sentence.
The main difference is that they use this to copy words into the output while our method uses it to influence the probabilities of all target words.%

\section{Constructing Lexicon Probabilities}
\label{sec:type}

In the previous section, we have defined some ways to use predictive probabilities $p_l(e_i | F, e_1^{i-1})$ based on word-to-word lexical probabilities $p_l(e | f)$.
Next, we define three ways to construct these lexical probabilities using automatically learned lexicons, handmade lexicons, or a combination of both.

\subsection{Automatically Learned Lexicons}
\label{sec:automatic}
In traditional SMT systems, lexical translation probabilities are generally learned directly from parallel data in an unsupervised fashion using a model such as the IBM models \cite{brown93cl,och03cl}.
These models can be used to estimate the alignments and lexical translation probabilities $p_l(e|f)$ between the tokens of the two languages using the expectation maximization (EM) algorithm.

First in the expectation step, the algorithm estimates the expected count $c(e | f)$.
In the maximization step, lexical probabilities are calculated by dividing the expected count by all possible counts: 
\begin{equation*}
    p_{l,a}(e|f) = \frac{c(f,e)}{\sum_{\tilde{e}} c(f,\tilde{e})},
\end{equation*}
The IBM models vary in level of refinement, with Model 1 relying solely on these lexical probabilities, and latter IBM models (Models 2, 3, 4, 5) introducing more sophisticated models of fertility and relative alignment. 
Even though IBM models also occasionally have problems when dealing with the rare words (e.g. ``garbage collecting'' effects \cite{liang06naacl}), traditional SMT systems generally achieve better translation accuracies of low-frequency words than NMT systems \cite{sutskever14nips}, indicating that these problems are less prominent than they are in NMT.

Note that in many cases, NMT limits the target vocabulary \cite{jean15acl} for training speed or memory constraints, resulting in rare words not being covered by the NMT vocabulary $V_E$.
Accordingly, we allocate the remaining probability assigned by the lexicon to the unknown word symbol $\langle \text{unk} \rangle$:
\begin{equation}
    \label{eq:unknown}
    p_{l,a}(e=\langle \text{unk} \rangle | f) = 1 - \sum_{i \in V_e} p_{l,a}(e=i|f).
\end{equation}

\subsection{Manual Lexicons}
\label{sec:handmade}
In addition, for many language pairs, broad-coverage handmade dictionaries exist, and it is desirable that we be able to use the information included in them as well.
Unlike automatically learned lexicons, however, handmade dictionaries generally do not contain translation probabilities.
To construct the probability $p_{l}(e | f)$, we define the set of translations $K_f$ existing in the dictionary for particular source word $f$, and assume a uniform distribution over these words:
\begin{equation*}
    p_{l,m}(e | f) = \begin {cases}
    \frac{1}{|K_f|} & \text{if } e \in K_f \\
    0             & \text{otherwise}
\end{cases}.
\end{equation*}
Following Equation (\ref{eq:unknown}), unknown source words will assign their probability mass to the $\langle \text{unk} \rangle$ tag.

\subsection{Hybrid Lexicons}
\label{sec:hybrid}

Handmade lexicons have broad coverage of words but their probabilities might not be as accurate as the learned ones, particularly if the automatic lexicon is constructed on in-domain data.
Thus, we also test a {\tt hybrid} method where we use the handmade lexicons to complement the automatically learned lexicon.\footnote{Alternatively, we could imagine a method where we combined the training data and dictionary before training the word alignments to create the lexicon. We attempted this, and results were comparable to or worse than the fill-up method, so we use the fill-up method for the remainder of the paper.}
\footnote{While most words in the $V_f$ will be covered by the learned lexicon, many words (13\% in experiments) are still left uncovered due to alignment failures or other factors.}
Specifically, inspired by phrase table fill-up used in PBMT systems \cite{bisazza11iwslt}, we use the probability of the automatically learned lexicons $p_{l,a}$ by default, and fall back to the handmade lexicons $p_{l,m}$ only for uncovered words:
\begin{equation}
    \label{eq:hybrid}
    p_{l,h}(e | f) = \begin{cases}
        p_{l,a}(e | f) & \text{if } f \text{ is covered}\\
        p_{l,m}(e | f) & \text{otherwise}
    \end{cases}
\end{equation}

\section{Experiment \& Result}
\label{sec:experiment}

In this section, we describe experiments we use to evaluate our proposed methods.

\subsection{Settings}

\noindent 
    {\bf Dataset:} We perform experiments on two widely-used tasks for the English-to-Japanese language pair: KFTT \cite{neubig11kftt} and BTEC \cite{kikui03interspeech}. 
    KFTT is a collection of Wikipedia article about city of Kyoto and BTEC is a travel conversation corpus.
    BTEC is an easier translation task than KFTT, because KFTT covers a broader domain, has a larger vocabulary of rare words, and has relatively long sentences.
    The details of each corpus are depicted in Table \ref{tab:data}.
    \begin{table}
        \centering
        \begin{tabular}{|l|r|r|r r|}
            \hline
            \multirow{2}{*}{Data} & \multirow{2}{*}{Corpus} &  \multirow{2}{*}{Sentence} & \multicolumn{2}{c|}{\underline{Tokens}} \\
                                  &        &                & \multicolumn{1}{c}{En}    & \multicolumn{1}{c|}{Ja} \\ \hline
            \multirow{2}{*}{Train}& BTEC   & 464K           & 3.60M & 4.97M \\
                                  & KFTT   & 377K           & 7.77M & 8.04M \\ \hline 
            \multirow{2}{*}{Dev}  & BTEC   & 510            & 3.8K & 5.3K \\ 
                                  & KFTT   & 1160           & 24.3K & 26.8K \\ \hline
            \multirow{2}{*}{Test} & BTEC   & 508            & 3.8K & 5.5K \\ 
                                  & KFTT   & 1169           & 26.0K & 28.4K \\ \hline
        \end{tabular}
        \caption{Corpus details.}
        \label{tab:data}
    \end{table}

\begin{table*}[t]
    \centering
    \begin{tabular}{|l| c c c|c c c|}
        \hline
        \multirow{2}{*}{System} & \multicolumn{3}{c|}{\underline{BTEC}} & \multicolumn{3}{c|}{\underline{KFTT}} \\ 
                                &  BLEU &NIST & RECALL & BLEU & NIST & RECALL  \\ \hline
         {\tt pbmt}             & 48.18 & 6.05 & 27.03 & 22.62 & 5.79 & 13.88 \\
         {\tt hiero}            & 52.27 & 6.34 & 24.32 &  22.54 & 5.82 & 12.83 \\ \hline
         {\tt attn}             & 48.31 & 5.98 & 17.39 &  20.86 & 5.15 & 17.68 \\
         {\tt auto-bias}        & \textbf{49.74}$^*$ & \textbf{6.11}$^*$  & \textbf{50.00}& \textbf{23.20}$^{\dag}$ & \textbf{5.59}$^{\dag}$ & \textbf{19.32} \\ 
         {\tt hyb-bias}         & \textbf{50.34}$^{\dag}$ & \textbf{6.10}$^*$ &  \textbf{41.67} & \textbf{22.80}$^{\dag}$ &  \textbf{5.55}$^{\dag}$ & 16.67 \\
        \hline   
     \end{tabular}
     \caption{
        Accuracies for the baseline attentional NMT ({\tt attn}) and the proposed bias-based method using the automatic ({\tt auto-bias}) or hybrid ({\tt hyb-bias}) dictionaries.
        Bold indicates a gain over the {\tt attn} baseline, $\dag$ indicates a significant increase at $p < 0.05$, and $*$ indicates $p < 0.10$.
        Traditional phrase-based ({\tt pbmt}) and hierarchical phrase based ({\tt hiero}) systems are shown for reference.
     }
     \label{tab:advertisement}
\end{table*}

    We tokenize English according to the Penn Treebank standard \cite{marcus93penntreebank} and lowercase, and tokenize Japanese using KyTea \cite{neubig11aclshort}.
    We limit training sentence length up to 50 in both experiments and keep the test data at the original length. 
    We replace words of frequency less than a threshold $u$ in both languages with the $\langle \text{unk} \rangle$ symbol and exclude them from our vocabulary.
    We choose $u=1$ for BTEC and $u=3$ for KFTT, resulting in $|V_f|=17.8$k, $|V_e|=21.8$k for BTEC and $|V_f|=48.2$k, $|V_e|=49.1$k for KFTT.

    \noindent
    {\bf NMT Systems:} We build the described models using the Chainer\footnote{\url{http://chainer.org/index.html}} toolkit.
    The depth of the stacking LSTM is $d=4$ and hidden node size $h=800$.
    We concatenate the forward and backward encodings (resulting in a 1600 dimension vector) and then perform a linear transformation to 800 dimensions.

    We train the system using the Adam \cite{kingma14corr} optimization method with the default settings: $\alpha=1\mathrm{e}{-3}, \beta_1=0.9, \beta_2=0.999, \epsilon=1\mathrm{e}{-8}$.
    Additionally, we add dropout \cite{srivastava14dropout} with drop rate $r=0.2$ at the last layer of each stacking LSTM unit to prevent overfitting. 
    We use a batch size of $B=64$ and we run a total of $N=14$ iterations for all data sets.
    All of the experiments are conducted on a single GeForce GTX TITAN X GPU with a 12 GB memory cache.

    At test time, we use beam search with beam size $b=5$.
    We follow \newcite{luong15acl} in replacing every unknown token at position $i$ with the target token that maximizes the probability $p_{l,a}(e_i | f_j)$. 
    We choose source word $f_j$ according to the highest alignment score in Equation (\ref{eq:attention}).
    This unknown word replacement is applied to both baseline and proposed systems.
    Finally, because NMT models tend to give higher probabilities to shorter sentences \cite{cho14ssst}, we discount the probability of $\langle \text{EOS} \rangle$ token by $10\%$ to correct for this bias. 

    \noindent
    {\bf Traditional SMT Systems:}
    We also prepare two traditional SMT systems for comparison: a PBMT system \cite{koehn03phrasebased} using Moses\footnote{\url{http://www.statmt.org/moses/}} \cite{koehn07acl}, and a hierarchical phrase-based MT system \cite{chiang07cl} using Travatar\footnote{\url{http://www.phontron.com/travatar/}} \cite{neubig13acldemo},
    Systems are built using the default settings, with models trained on the training data, and weights tuned on the development data.

    \noindent
    {\bf Lexicons:}
    We use a total of 3 lexicons for the proposed method, and apply {\tt bias} and {\tt linear} method for all of them, totaling 6 experiments.
    The first lexicon ({\tt auto}) is built on the training data using the automatically learned lexicon method of \S\ref{sec:automatic} separately for both the BTEC and KFTT experiments.
    Automatic alignment is performed using GIZA++ \cite{och03cl}.
    The second lexicon ({\tt man}) is built using the popular English-Japanese dictionary Eijiro\footnote{\url{http://eijiro.jp}} with the manual lexicon method of \S\ref{sec:handmade}.
    Eijiro contains 104K distinct word-to-word translation entries.
    The third lexicon ({\tt hyb}) is built by combining the first and second lexicon with the hybrid method of \S\ref{sec:hybrid}.
    
    \noindent
    {\bf Evaluation:}
    We use standard single reference BLEU-4 \cite{papineni02acl} to evaluate the translation performance. 
    Additionally, we also use NIST \cite{doddington02hlt}, which is a measure that puts a particular focus on low-frequency word strings, and thus is sensitive to the low-frequency words we are focusing on in this paper.
    We measure the statistical significant differences between systems using paired bootstrap resampling \cite{koehn04emnlp} with 10,000 iterations and measure statistical significance at the $p<0.05$ and $p<0.10$ levels.

    Additionally, we also calculate the recall of rare words from the references.
    We define ``rare words'' as words that appear less than eight times in the target training corpus or references, and measure the percentage of time they are recovered by each translation system.

\subsection{Effect of Integrating Lexicons}

\begin{figure}[t]
\centering
    \includegraphics[width=\linewidth]{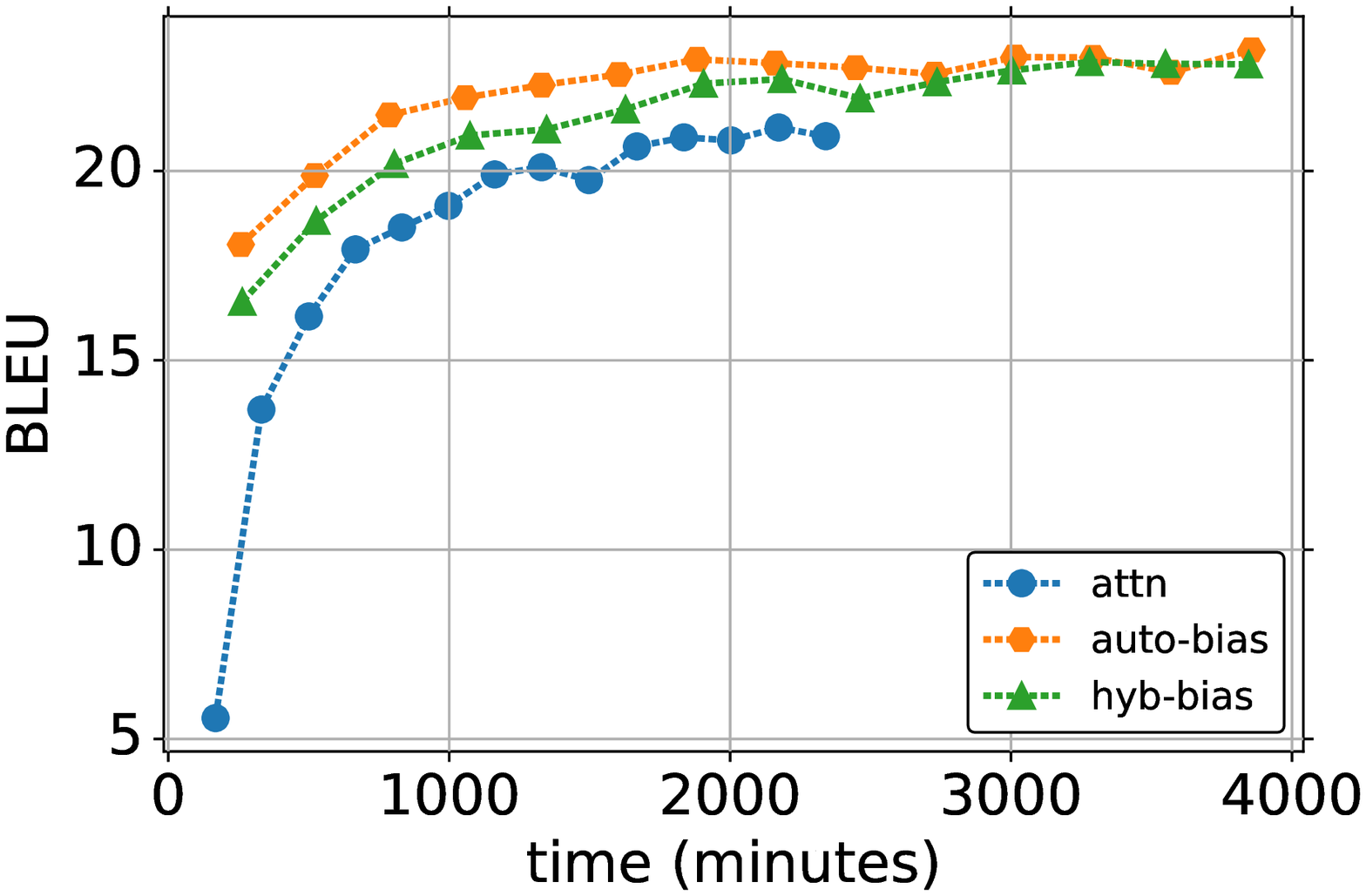}
    \label{fig:kftt_bleu}
\caption{Training curves for the baseline {\tt attn} and the proposed {\tt bias} method.}
\label{fig:full}
\end{figure}

In this section, we first a detailed examination of the utility of the proposed {\tt bias} method when used with the {\tt auto} or {\tt hyb} lexicons, which empirically gave the best results, and perform a comparison among the other lexicon integration methods in the following section.
Table \ref{tab:advertisement} shows the results of these methods, along with the corresponding baselines.

\begin{table*}[ht!]
\centering
    \begin{tabular}{|l|l|}
        \hline
        Input              & Do you have an opinion regarding \underline{extramarital affairs}? \\
        Reference             & \underline{不倫} に 関して 意見 が あります か。 \\
                           & \small{\underline{Furin} ni kanshite iken ga arimasu ka.} \\
        \texttt{attn}      & \underline{サッカー} に 関する 意見 は あります か。 \\
                           & \small{\underline{Sakk\={a}} ni kansuru iken wa arimasu ka.} \textit{(Do you have an opinion about soccer?)} \\
        \texttt{auto-bias} & \underline{不倫} に 関して 意見 が あります か。 \\
                           & \small{\underline{Furin} ni kanshite iken ga arimasu ka.} \textit{(Do you have an opinion about affairs?)} \\
        \hline
        Input              & Could you put these \underline{fragile things} in a safe place?\\
        Reference          & この \underline{壊れ物} を 安全な 場所 に 置いて もらえません か 。\\
                           & \small{Kono \underline{kowaremono} o anzen'na basho ni oite moraemasen ka.} \\
        \texttt{attn}      & \underline{貴重品} を 安全 に 出したい の ですが 。\\
                           & \small{\underline{Kich\={o}-hin} o anzen ni dashitai nodesuga.} \textit{(I'd like to safely put out these valuables.)} \\
        \texttt{auto-bias} & この \underline{壊れ物} を 安全な 場所 に 置いて もらえません か 。\\
                           & \small{Kono \underline{kowaremono} o anzen'na basho ni oite moraemasen ka.} \\
                           & \small{\textit{(Could you put these fragile things in a safe place?)}} \\ 
        \hline
    \end{tabular}
    \caption{Examples where the proposed \texttt{auto-bias} improved over the baseline system \texttt{attn}. Underlines indicate words were mistaken in the baseline output but correct in the proposed model's output.}
   \label{tab:res_ex}
\end{table*}

First, compared to the baseline {\tt attn}, our {\tt bias} method achieved consistently higher scores on both test sets.
In particular, the gains on the more difficult KFTT set are large, up to 2.3 BLEU, 0.44 NIST, and 30\% Recall, demonstrating the utility of the proposed method in the face of more diverse content and fewer high-frequency words.

Compared to the traditional {\tt pbmt} systems {\tt hiero}, particularly on KFTT we can see that the proposed method allows the NMT system to exceed the traditional SMT methods in BLEU.
This is despite the fact that we are not performing ensembling, which has proven to be essential to exceed traditional systems in several previous works \cite{sutskever14nips,luong15emnlp,sennrich16acl}. 
Interestingly, despite gains in BLEU, the NMT methods still fall behind in NIST score on the KFTT data set, demonstrating that traditional SMT systems still tend to have a small advantage in translating lower-frequency words, despite the gains made by the proposed method.

In Table \ref{tab:res_ex}, we show some illustrative examples where the proposed method (\texttt{auto-bias}) was able to obtain a correct translation while the normal attentional model was not.
The first example is a mistake in translating ``extramarital affairs'' into the Japanese equivalent of ``soccer,'' entirely changing the main topic of the sentence.
This is typical of the errors that we have observed NMT systems make (the mistake from Figure \ref{fig:intro_mistake} is also from \texttt{attn}, and was fixed by our proposed method).
The second example demonstrates how these mistakes can then affect the process of choosing the remaining words, propagating the error through the whole sentence.

Next, we examine the effect of the proposed method on the training time for each neural MT method, drawing training curves for the KFTT data in Figure \ref{fig:full}.
Here we can see that the proposed \texttt{bias} training methods achieve reasonable BLEU scores in the upper 10s even after the first iteration.
In contrast, the baseline \texttt{attn} method has a BLEU score of around 5 after the first iteration, and takes significantly longer to approach values close to its maximal accuracy.
This shows that by incorporating lexical probabilities, we can effectively bootstrap the learning of the NMT system, allowing it to approach an appropriate answer in a more timely fashion.%
\footnote{Note that these gains are despite the fact that one iteration of the proposed method takes a longer (167 minutes for \texttt{attn} vs. 275 minutes for \texttt{auto-bias}) due to the necessity to calculate and use the lexical probability matrix for each sentence. It also takes an additional 297 minutes to train the lexicon with GIZA++, but this can be greatly reduced with more efficient training methods \cite{dyer13naacl}.}

\begin{figure}[t]
    \centering
    \includegraphics[width=\linewidth]{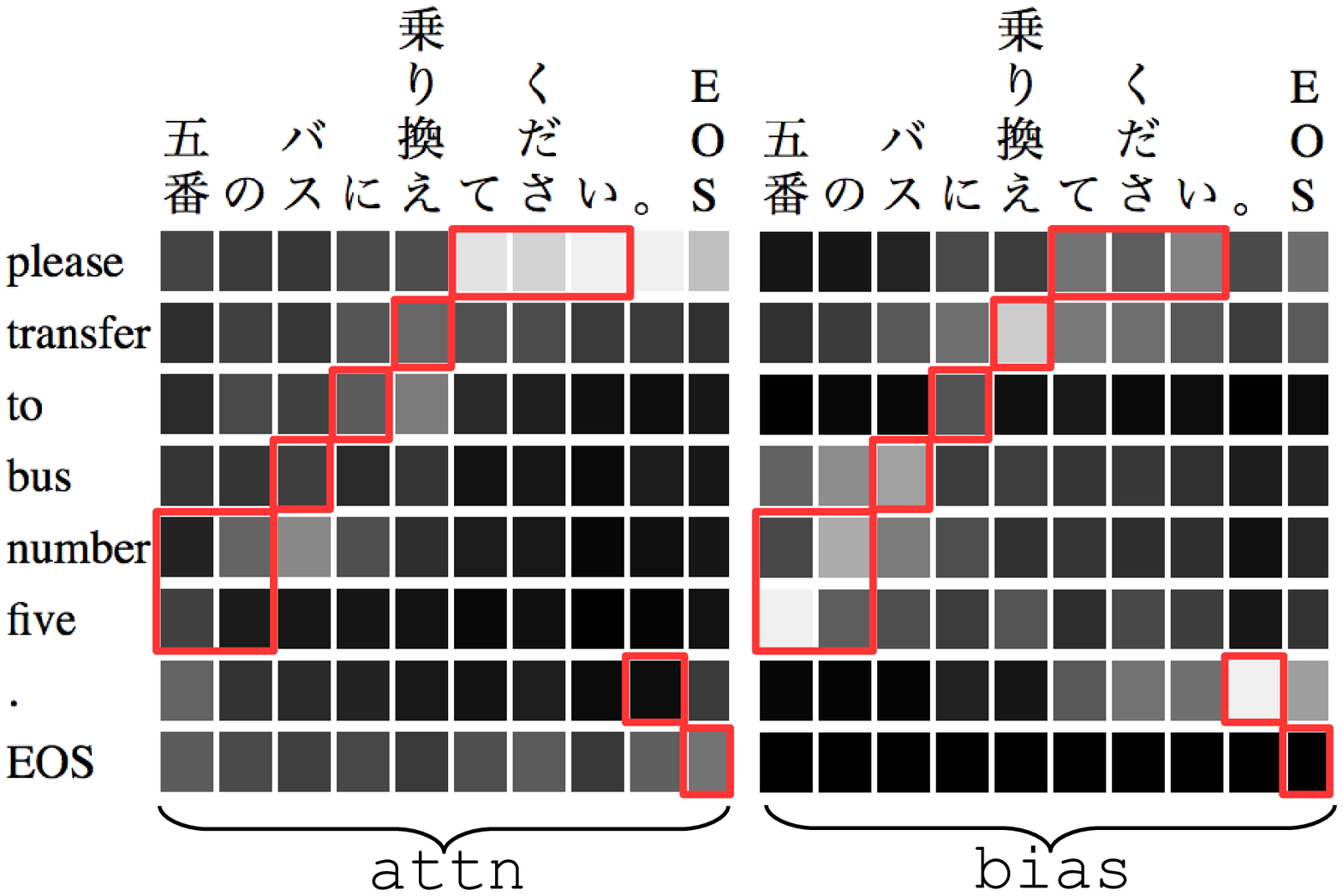}
    \caption{Attention matrices for baseline \texttt{attn} and proposed \texttt{bias} methods.
    Lighter colors indicate stronger attention between the words, and boxes surrounding words indicate the correct alignments.
    }
    \label{fig:attention}
\end{figure}

It is also interesting to examine the alignment vectors produced by the baseline and proposed methods, a visualization of which we show in Figure \ref{fig:attention}.
For this sentence, the outputs of both methods were both identical and correct, but we can see that the proposed method (right) placed sharper attention on the actual source word corresponding to content words in the target sentence.
This trend of peakier attention distributions in the proposed method held throughout the corpus, with the per-word entropy of the attention vectors being 3.23 bits for \texttt{auto-bias}, compared with 3.81 bits for \texttt{attn}, indicating that the \texttt{auto-bias} method places more certainty in its attention decisions.



\subsection{Comparison of Integration Methods}

\begin{table}[t]
\centering
\begin{subtable}[t]{\linewidth}
    \caption{BTEC}
    \begin{tabular}{|l| c c|c c |}
        \hline
         \multirow{2}{*}{Lexicon}      & \multicolumn{2}{c|}{\underline{BLEU}} & \multicolumn{2}{c|}{\underline{NIST}} \\ 
                                  & {\tt bias} & {\tt linear} & {\tt bias} & {\tt linear} \\ \hline
         -                        & \multicolumn{2}{c|}{48.31} & \multicolumn{2}{c|}{5.98} \\ \hline
         {\tt auto}               & \textbf{49.74}$^*$ & 47.97~ &  \textbf{6.11} & 5.90~ \\
         {\tt man}                & \textbf{49.08}~ & \textbf{51.04}$^\dag$ & \textbf{6.03}$^*$&  \textbf{6.14}$^\dag$  \\ 
         {\tt hyb}                & \textbf{50.34}$^\dag$ & \textbf{49.27}~ & \textbf{6.10}$^*$ & 5.94~  \\
        \hline
        
     \end{tabular}
    \label{tab:bias}
\end{subtable}
\begin{subtable}[t]{\linewidth}
    \caption{KFTT}
    \begin{tabular}{|l| c c|c c |}
        \hline
         \multirow{2}{*}{Lexicon}      & \multicolumn{2}{c|}{\underline{BLEU}} & \multicolumn{2}{c|}{\underline{NIST}} \\ 
                                  & {\tt bias} & {\tt linear} & {\tt bias} & {\tt linear} \\ \hline
         -                        & \multicolumn{2}{c|}{20.86} & \multicolumn{2}{c|}{5.15} \\ \hline
         {\tt auto}               & \textbf{23.20}$^\dag$ &  18.19~  &  \textbf{5.59}$^\dag$ & 4.61~ \\
         {\tt man}                & 20.78~ & \textbf{20.88}~ &  5.12~ &  5.11~\\ 
         {\tt hyb}                & \textbf{22.80}$^\dag$ & 20.33~ &  \textbf{5.55}$^\dag$ & 5.03~\\
        \hline
     \end{tabular}
     \label{tab:linear}
\end{subtable}
\caption{
  A comparison of the \texttt{bias} and \texttt{linear} lexicon integration methods on the automatic, manual, and hybrid lexicons.
  The first line without lexicon is the traditional attentional NMT.
        }   
\label{tab:full}
\end{table}

Finally, we perform a full comparison between the various methods for integrating lexicons into the translation process, with results shown in Table \ref{tab:full}.
In general the {\tt bias} method improves accuracy for the {\tt auto} and {\tt hyb} lexicon, but is less effective for the {\tt man} lexicon.
This is likely due to the fact that the manual lexicon, despite having broad coverage, did not sufficiently cover target-domain words (coverage of unique words in the source vocabulary was 35.3\% and 9.7\% for BTEC and KFTT respectively).

Interestingly, the trend is reversed for the {\tt linear} method, with it improving {\tt man} systems, but causing decreases when using the {\tt auto} and {\tt hyb} lexicons.
This indicates that the \texttt{linear} method is more suited for cases where the lexicon does not closely match the target domain, and plays a more complementary role.
Compared to the log-linear modeling of \texttt{bias}, which strictly enforces constraints imposed by the lexicon distribution \cite{klakow98icslp}, linear interpolation is intuitively more appropriate for integrating this type of complimentary information.

On the other hand, the performance of linear interpolation was generally lower than that of the bias method.
One potential reason for this is the fact that we use a constant interpolation coefficient that was set fixed in every context.
\newcite{gu16acl} have recently developed methods to use the context information from the decoder to calculate the different interpolation coefficients for every decoding step, and it is possible that introducing these methods would improve our results.

\section{Additional Experiments}


To test whether the proposed method is useful on larger data sets, we also performed follow-up experiments on the larger Japanese-English ASPEC dataset \cite{nakazawa16lrec} that consist of 2 million training examples, 63 million tokens, and 81,000 vocabulary size.
We gained an improvement in BLEU score from 20.82 using the \texttt{attn} baseline to 22.66 using the \texttt{auto-bias} proposed method.
This experiment shows that our method scales to larger datasets.

\section{Related Work}
\label{sec:related}

From the beginning of work on NMT, unknown words that do not exist in the system vocabulary have been focused on as a weakness of these systems.
Early methods to handle these unknown words replaced them with appropriate words in the target vocabulary \cite{jean15acl,luong15acl} according to a lexicon similar to the one used in this work.
In contrast to our work, these only handle unknown words and do not incorporate information from the lexicon in the learning procedure.

There have also been other approaches that incorporate models that learn when to copy words as-is into the target language \cite{allamanis16icml,gu16acl,gulcehre16acl}.
These models are similar to the \texttt{linear} approach of \S\ref{sec:linear}, but are only applicable to words that can be copied as-is into the target language.
In fact, these models can be thought of as a subclass of the proposed approach that use a lexicon that assigns a all its probability to target words that are the same as the source.
On the other hand, while we are simply using a static interpolation coefficient $\lambda$, these works generally have a more sophisticated method for choosing the interpolation between the standard and ``copy'' models.
Incorporating these into our \texttt{linear} method is a promising avenue for future work.

In addition \newcite{mi16acl} have also recently proposed a similar approach by limiting the number of vocabulary being predicted by each batch or sentence.
This vocabulary is made by considering the original HMM alignments gathered from the training corpus.
Basically, this method is a specific version of our bias method that gives some of the vocabulary a bias of negative infinity and all other vocabulary a uniform distribution.
Our method improves over this by considering actual translation probabilities, and also considering the attention vector when deciding how to combine these probabilities.

Finally, there have been a number of recent works that improve accuracy of low-frequency words using character-based translation models \cite{ling15corr,costajussa16acl,chung16acl}.
However, \newcite{luong16acl} have found that even when using character-based models, incorporating information about words allows for gains in translation accuracy, and it is likely that our lexicon-based method could result in improvements in these hybrid systems as well.

\section{Conclusion \& Future Work}
\label{sec:conclusion}
In this paper, we have proposed a method to incorporate discrete probabilistic lexicons into NMT systems to solve the difficulties that NMT systems have demonstrated with low-frequency words.
As a result, we achieved substantial increases in BLEU (2.0-2.3) and NIST (0.13-0.44) scores, and observed qualitative improvements in the translations of content words.

For future work, we are interested in conducting the experiments on larger-scale translation tasks.
We also plan to do subjective evaluation, as we expect that improvements in content word translation are critical to subjective impressions of translation results.
Finally, we are also interested in improvements to the \texttt{linear} method where $\lambda$ is calculated based on the context, instead of using a fixed value.

\section*{Acknowledgment}
We thank Makoto Morishita and Yusuke Oda for their help in this project. 
We also thank the faculty members of AHC lab for their supports and suggestions.

This work was supported by grants from the Ministry of Education, Culture, Sport, Science, and Technology of Japan and in part by JSPS KAKENHI Grant Number 16H05873.

\bibliographystyle{emnlp2016}
\bibliography{myabbrv,emnlp2016}

\end{CJK}
\end{document}